\documentclass[10pt,twocolumn]{article}

\usepackage[margin=0.75in]{geometry}
\usepackage{graphicx}
\usepackage{amsmath,amssymb}
\usepackage{hyperref}
\usepackage{url}
\usepackage{booktabs}
\usepackage{multirow}
\usepackage{xcolor}
\usepackage{microtype}
\usepackage{tikz}
\usetikzlibrary{positioning, calc, arrows.meta, shapes.geometric, fit, backgrounds}
\usepackage{subcaption}
\usepackage{enumitem}
\usepackage{natbib}
\usepackage{pgfplots}
\pgfplotsset{compat=1.18}

\hypersetup{
  pdftitle={Operationalizing Document AI: A Microservice Architecture for OCR and LLM Pipelines in Production},
  pdfkeywords={Document AI, Document Understanding, Microservices, OCR, LLM, Production Systems, Scalability},
  colorlinks=true,
  linkcolor={blue!70!black},
  citecolor={blue!70!black},
  urlcolor={blue!70!black}
}

\title{\textbf{Operationalizing Document AI: A Microservice Architecture for OCR and LLM Pipelines in Production}}

\author{
Yao Fehlis, Benjamin Bengfort, Zhangzhang Si, Vahid Eyorokon, Prema Roman, Patrick Deziel, \\
Devon Slonaker, Steve Veldman, Ben Johnson, Joyce Rigelo, Michael Wharton, Steve Kramer \\[2pt]
\textsc{Kungfu.ai}
}

\date{}

\begin{document}
\maketitle

\begin{abstract}
Academic research tends to focus on new models for document understanding creating a wide gap in the literature between model definition and running models at production scale. To close that gap, we present a microservice architecture that encapsulates pipelines of multiple models for classification, optical character recognition (OCR), and large language model structured field extraction as well as our experience running this pipeline on thousands of multi-page documents per hour. We describe our primary design decisions, including a hybrid classification, separation of GPU-bound inference from CPU-bound orchestration, use of asynchronous processing for the many IO-bound operations in the pipeline, and an independent, horizontal scaling strategy. Using batch profiling, we identified two surprising qualitative findings that shape production deployments: OCR, not language-model parsing, dominates end-to-end latency, and the system saturates at a concurrency determined by shared GPU-inference capacity rather than worker count. Our goal is to provide practitioners with concrete architectural patterns for building document understanding systems that work beyond the benchmark; effectively operationalizing models in production.
\end{abstract}

\section{Introduction}\label{sec:introduction}

The document understanding research community produces a steady stream of model innovations---LayoutLM~\citep{xu2020layoutlm}, DocTR~\citep{mindee2021doctr}, Donut~\citep{kim2022donut}, Pix2Struct~\citep{lee2023pix2struct}, and dozens of vision-language models (VLMs)---each advancing accuracy on benchmarks like DocVQA~\citep{mathew2021docvqa} and FUNSD~\citep{jaume2019funsd}. Yet a practitioner attempting to deploy these models into a production system that processes thousands of forms per day finds little guidance on the engineering required to make them work reliably.

The gap between a model checkpoint and on-demand model usage is substantial. Models must be containerized and served behind inference APIs. Documents arrive in heterogeneous formats---multi-page TIFFs, scanned PDFs, photographs---and must be normalized before processing often using GPU-accelerated operations. Classification must route each document to the correct extraction pipeline. OCR must handle scanning artifacts, skewed pages, and degraded image quality. LLMs must be used to extract fields that must be validated and structured correctly. All of these operations must effectively wrap multiple model compute paradigms that are generally implemented efficiently for batch processing (e.g. training and multi-instance inference) but not for dynamic switching between model types while balancing CPU-, GPU-, and IO-bound operatations with fixed memory resources.

Solving on-demand model usage is only the first step: moving from model usage to a production system requires addressing a wide range of other challenges from managing the differences between failures and errors to determining the configuration that best suits both the model and the workload. Other considerations such as implementing timeouts and retries to handle stochastic failures, outputing model meta information such as confidence scores, and model operation tracing for explainability blur the line between scientific research and software engineering. For deployment scenarios that involve sensitive, private, and/or proprietary information, the system architecture must support processing entirely within a secure cloud enclave and/or with an on-premise computing platform with sufficient scanning, version control, logging, and access controls. All of this must happen at scale, with observable failures, predictable latency, and, of course, \emph{manageable cost}.

This paper describes an architecture for scalable and fault-tolerant production system for structured form processing that addresses these challenges. We also describe a system that we built using this architecture, which processes scanned, multi-page documents through a pipeline of classification, OCR, text stitching, and language-model-based field extraction. Finally we describe our experiences deploying the architecture as three microservices on Kubernetes with message queue-based work distribution and object storage for documents and how we reduced the cost of proessing documents from \$0.01 per page to \$0.001 per page while maintaining a 96\% accuracy.

We make three contributions. First, we describe the architecture and the design decisions behind it, including why we decompose the system into three services rather than a monolith, and how we separate GPU-bound inference from CPU-bound orchestration (\S\ref{sec:architecture}). Second, we walk through the pipeline design in detail, including a hybrid classification strategy that balances cost and accuracy, and the flow from raw images to structured JSON output (\S\ref{sec:pipeline}). Third, we report qualitative findings from batch profiling that reveal the system's scaling behavior and bottlenecks, together with lessons learned from production operation (\S\ref{sec:case-study}).

\section{Related Work}\label{sec:related}

\paragraph{Document understanding models.}
The model landscape spans traditional OCR (Tesseract~\citep{smith2007tesseract}, PaddleOCR~\citep{du2020ppocr}), layout-aware transformers (LayoutLM~\citep{xu2020layoutlm}, LayoutLMv3~\citep{huang2022layoutlmv3}), end-to-end image-to-text models (Donut~\citep{kim2022donut}, Nougat~\citep{blecher2023nougat}), general-purpose VLMs~\citep{anthropic2024claude,google2024gemini,bai2023qwenvl}, hybrid OCR-into-VLM designs~\citep{nacson2024docvlm}, and CPU-optimized open models such as Docling~\citep{ibm2024docling}. Each advances accuracy on benchmarks but leaves open how to compose these components into a production system.

\paragraph{Production ML systems.}
System-level descriptions of production ML deployments remain less common than model papers. TFX~\citep{baylor2017tfx} describes Google's end-to-end ML platform. Uber's Michelangelo~\citep{hermann2017uber} and Facebook's FBLearner~\citep{dunn2016fblearner} address ML workflow management. For document processing specifically, enterprise platforms like ABBYY, Kofax, and cloud services (AWS Textract, Google Document AI, Azure Form Recognizer) exist as commercial offerings, but their architectures are proprietary and undocumented in the literature. Recent work has begun to describe enterprise and multimodal document-processing systems: IDP Accelerator~\citep{islam2026idp} presents an agentic document-intelligence framework with document splitting, extraction, analytics, and compliance validation; MMORE~\citep{sallinen2025mmore} describes a modular distributed pipeline for multimodal retrieval-augmented generation and extraction across heterogeneous file types; and domain-specific systems combine OCR, classifiers, and VLMs for claims or copy-heavy enterprise extraction~\citep{cheng2026hybridclaims,wang2025hybridocrllm}. These systems show that large-scale document AI is an active area; our focus is the service-level architecture and operational lessons for a structured-form extraction pipeline.

\paragraph{OCR versus multimodal extraction.}
OCR-free and image-native document models such as Donut~\citep{kim2022donut} motivate simpler pipelines that send page images directly to a VLM. Recent benchmarking suggests that powerful multimodal LLMs may match OCR-enhanced approaches for some business-document extraction tasks~\citep{shen2026ocrornot}. Our system uses OCR-first extraction for cost control, auditability, page-level intermediate artifacts, and compatibility with text-only parsing models, but the architecture treats this as a configurable tradeoff rather than a permanent modeling assumption. We refer readers to our prior practical guide~\citep{fehlis2025docunderstanding} for a capability-dimension framework (text recognition, structural understanding, output flexibility, spatial awareness, task adaptability) that can inform this OCR-versus-VLM choice on a per-deployment basis.

\paragraph{Document retrieval.}
ColPali~\citep{faysse2024colpali} introduces late-interaction embeddings for document retrieval, enabling efficient page-level search without OCR. This retrieval-augmented paradigm complements extraction pipelines like ours by enabling selective processing of relevant pages from large document collections.

\section{System Architecture}\label{sec:architecture}

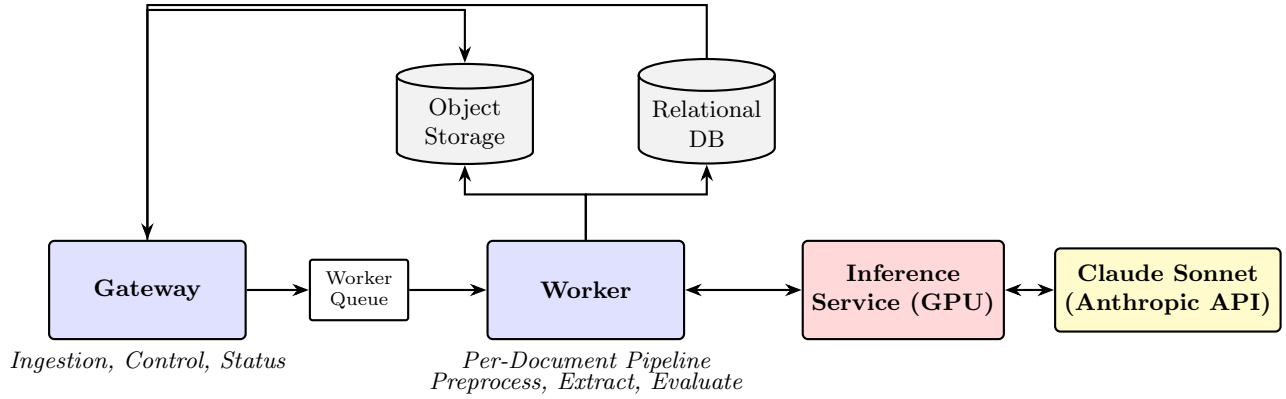
\begin{figure*}[t]
\centering
\begin{tikzpicture}[
    svc/.style={draw, thick, rounded corners=2pt, align=center, minimum width=2.6cm, minimum height=1.3cm, font=\small\bfseries, fill=blue!12},
    gpu/.style={draw, thick, rounded corners=2pt, align=center, minimum width=2.6cm, minimum height=1.3cm, font=\small\bfseries, fill=red!15},
    ext/.style={draw, thick, rounded corners=2pt, align=center, minimum width=2.2cm, minimum height=1.1cm, font=\small\bfseries, fill=yellow!25},
    store/.style={draw, thick, cylinder, shape border rotate=90, aspect=0.25, align=center, minimum width=1.8cm, minimum height=1.2cm, font=\small, fill=gray!10},
    q/.style={draw, thick, rounded corners=1pt, align=center, minimum width=1.3cm, minimum height=0.8cm, font=\scriptsize, fill=white},
    sub/.style={font=\scriptsize\itshape, align=center},
    arr/.style={-{Stealth[length=2.2mm]}, thick},
    barr/.style={{Stealth[length=2.2mm]}-{Stealth[length=2.2mm]}, thick}
]

\node[store] (s3)  at (0, 2.2)  {Object\\Storage};
\node[store] (pg)  at (3.2, 2.2){Relational\\DB};

\node[svc] (gw)  at (-4.2, 0) {Gateway};
\node[sub, below=0.02cm of gw] {\small Ingestion, Control, Status};

\node[q]   (message-queue) at (-1.4, 0)  {Worker\\Queue};

\node[svc] (wrk) at (1.6, 0)   {Worker};
\node[sub, below=0.02cm of wrk] {\small Per-Document Pipeline\\\small Preprocess, Extract, Evaluate};

\node[gpu] (inf) at (5.8, 0)   {Inference\\Service (GPU)};

\node[ext] (vlm) at (9.3, 0)   {Claude Sonnet\\(Anthropic API)};

\draw[arr] (gw.east) -- (message-queue.west);
\draw[arr] (message-queue.east) -- (wrk.west);
\draw[barr] (wrk.east) -- (inf.west);
\draw[barr] (inf.east) -- (vlm.west);

\draw[arr] (wrk.north) -- ++(0,0.6) -| (s3.south);
\draw[arr] (wrk.north) -- ++(0,0.6) -| (pg.south);

\draw[arr] (gw.north) |- ($ (s3.north) + (0,0.7) $) -| (s3.north);
\draw[arr] (pg.north) |- ($ (pg.north) + (0, 0.7) $) -| (gw.north);

\end{tikzpicture}
\caption{System architecture. The Gateway accepts submissions, persists page images in object storage and tracking records in the relational DB, and enqueues document IDs onto a message queue. Workers pull documents, perform CPU-bound orchestration, and call the Inference Service for GPU-bound OCR and the Anthropic API (Claude Sonnet) for VLM-based steps and language-model parsing. Separating CPU-bound orchestration from GPU-bound inference enables each tier to scale independently.}
\label{fig:architecture}
\end{figure*}

The architecture decomposes document understanding into three microservices, each independently deployable and scalable (Figure~\ref{fig:architecture}). This decomposition reflects a fundamental insight: the computational profiles of inference (GPU-bound, high memory) and orchestration (CPU-bound, I/O-heavy) differ enough that coupling them wastes resources and limits scaling flexibility.

By decoupling inference, CPU-bound services primarily utilize I/O-heavy operations required for document processing, allowing them to take advantage of asynchronous coroutines without the need for internal parallelization. Although this approach means that inference services cannot be scaled using batch-processing without buffering, it does allow for independent scaling of each service. This architecture is therefore well-suited to tuning to specific workloads without over-provisioning resources that would otherwise sit idle during I/O operations.

\subsection{Gateway: Ingestion Service}\label{sec:gateway}

The Gateway service is the system's entry and exit point. It accepts document submissions through two inbound paths: a REST API for synchronous submissions from clients and an ingestion queue for asynchronous handoff from upstream pipelines. The Gateway is responsible for storing page images in object storage, creating tracking records in a relational database, and enqueuing document IDs to notify workers the document is ready for processing. It also serves a web-based inspection UI for human review of extraction results and reports processing status via a status queue.

The Gateway is almost entirely I/O-bound and must scale to the expected rate of ingestion throughput. For example, commercial scanners are capable of scanning 150 pages per minute~\citep{ricohfi8950} at 300 dpi producing page images of 2-90 MiB requiring a bandwidth averaging 225 MiB/s. For physical document scanning workloads, this bandwidth is "bursty" as scanning throughput is limited by the size of the document feeder and normal work hours for the facility. In this simple example, it is easy to see that ingestion throughput is often fundamentally different than the throughput of the rest of the system, requiring a different scaling strategy from other document processing workloads.

Because of the difference in scaling strategy, the Gateway is intentionally lightweight---it performs no inference or extraction, nor any other CPU intensive operations. This keeps its resource footprint small (tuned to ingestion) and its failure modes simple: if the Gateway is down, no new documents enter the system, but in-flight processing continues unaffected.

\subsection{Worker: Pipeline Orchestration}\label{sec:worker}

Workers serve as the pipeline's execution engine. They orchestrate the entire workflow and run individual steps as dictated by the system configuration. Each worker pod runs multiple concurrent tasks (limited to a maximum number to prevent resource contention), pulling document IDs from the message queue, and executing the configured extraction pipeline. Workers download page images from object storage on demand (lazy loading), invoke the Inference Service for inference-heavy steps, and upload structured JSON results upon completion.

Workers are CPU-bound and I/O-heavy: they spend most of their time waiting for inference responses, database reads and writes, downloading and uploading images, and marshaling data between pipeline steps. We therefore use asynchronous task execution inside each worker process so one task can make progress while another waits on inference or I/O. This provides vertical concurrency within a pod, while Kubernetes provides horizontal scaling by adding more worker pods. Throughput increases with worker concurrency until the Inference Service or downstream APIs saturate.

The effective concurrency of the system is $\text{pods} \times \text{tasks per pod}$. With the default configuration of 5 tasks per pod, a deployment of 5 worker pods provides 25 concurrent document processing slots.

\subsection{Inference Service}\label{sec:inference}

The Inference Service isolates GPU-bound inference behind a REST API, exposing OCR models (e.g., DocTR~\citep{mindee2021doctr}) and VLM capabilities (via cloud API proxying to managed services) to workers. This separation provides three benefits:

\begin{enumerate}[nosep]
\item \textbf{Independent scaling.} GPU nodes are expensive. Decoupling inference from orchestration means we provision GPU capacity based on inference demand, not worker count.
\item \textbf{Model swapping.} New OCR models can be deployed to the Inference Service without touching worker code. For example, we have swapped among DocTR, Docling~\citep{ibm2024docling}, and SmolDocling with configuration changes only.
\item \textbf{Resource isolation.} DocTR requires $\sim$800\,MB of GPU memory. Running it inside worker pods would either waste GPU resources (because most worker time is non-inference) or force CPU-only inference (3--5$\times$ slower).
\end{enumerate}

An important consideration for the deployment of inference service(s) is ensuring computational \textit{staggering}. Processing a single document is a sequential process where each step follows from the previous step. However, document processing steps can be executed in parallel---GPU operations can be performed during CPU operations, and multi-core CPUs can effectively handle multiple steps. Staggering multiple sequential processes allows this parallelism and to saturate the compute resources available if the steps are carefully designed to provide this advantage.

To achieve staggering, the inference service must use hybrid capabilities. At first, we were concerned primarily with VLM throughput and batching. However, by utilizing different models for classification, OCR, data extraction, validation, and more, we were able to achieve staggering which led to better resource utilization (and better performance for cost) than batching alone.

\subsection{Queue-Driven Communication}\label{sec:queues}

Asynchronous job coordination flows through message queues, providing backpressure, retry semantics, and decoupled scaling. Inference steps use direct request-response service APIs; queues coordinate document ownership, status propagation, and retry behavior. Three queues serve distinct roles:

\begin{itemize}[nosep]
\item \textbf{Ingestion queue}: External systems submit document IDs for processing.
\item \textbf{Worker queue}: The Gateway enqueues validated documents; workers consume them.
\item \textbf{Status queue}: Workers publish completion notifications; the Gateway consumes them for status reporting.
\end{itemize}

Queue-based communication means any service can be restarted or scaled without affecting others. If workers fall behind, messages accumulate in the worker queue, providing natural backpressure rather than cascading failures. Queues also coordinate the document processing steps, ensuring that only one worker processes a document at a time, ensuring that the state of the document is modified in an idempotent manner during processing even when a failure occurs. This guarantee allows us to use checkpointing and retry mechanisms that do not require reprocessing the entire document from the beginning.

\subsection{Failure Isolation and Service Contracts}\label{sec:contracts}

The microservice boundary is useful not only for scaling but also for failure isolation. The Gateway, Workers, and Inference Service communicate through narrow contracts: object storage is the source of truth for page artifacts, queue messages carry lightweight work references rather than full payloads, and the Inference Service exposes stable request-response interfaces for OCR and model-backed extraction. This approach keeps orchestration logic independent of model implementation details and prevents transient model-serving failures from propagating arbitrary state back into the rest of the system.

These contracts also define restart behavior. If the Gateway becomes unavailable, no new work enters the system, but in-flight documents continue processing because workers already hold queue leases and page artifacts remain in object storage. At each step, data is checkpointed and the processing status is updated in the database, ensuring that if a step fails, processing can resume from the point of failure with checkpointed data reloaded into memory. If workers restart, the queue redelivery mechanism provides recovery without requiring model state to be reconstructed inside the orchestration tier. If the Inference Service is unavailable or saturated, workers block or retry at a well-defined service boundary rather than failing the entire control plane. In practice, this isolation makes operational debugging substantially easier because failures can be localized to ingress, orchestration, inference, or downstream systems instead of appearing as a single monolithic outage.

\subsection{When This Architecture Is Appropriate}\label{sec:when-to-use}

This architecture is most useful when document volume, page count, or model cost is high enough that per-step scaling and cost control matter. It is also useful when teams need control over model selection, data residency, intermediate artifacts, or human-review workflows. For low-volume, homogeneous forms, a managed document-AI platform or a simpler monolithic worker may be cheaper to operate.

We considered four common alternatives. A managed intelligent-document-processing platform can reduce engineering effort, but bespoke, dataset-specific pipelines typically achieve substantially higher extraction accuracy on the document types that matter to a given deployment---managed platforms also limit model choice, observability into intermediate artifacts, and deployment control. A monolithic worker that performs ingestion, OCR, parsing, and result delivery is simpler, but couples CPU orchestration to GPU or API inference capacity. Embedding OCR models directly inside every worker removes a network call, but its co-located resource profile is poor in both directions: GPU-backed workers idle their accelerators during the majority of each task that is spent on I/O and orchestration (DocTR alone reserves $\sim$800\,MB of GPU memory per worker, \S\ref{sec:inference}), while CPU-only workers run OCR 3--5$\times$ slower and bottleneck the pipeline. A VLM-only pipeline is attractive because it bypasses OCR and text stitching, but it usually increases per-page cost, complicates auditability, and makes it harder to preserve word-level evidence. The proposed design accepts more infrastructure complexity in exchange for independent scaling, model replaceability, and clearer operational boundaries.

\section{Pipeline Design}\label{sec:pipeline}

Each document flows through a configurable sequence of pipeline steps, defined per document type in a YAML configuration. The pipeline implements a modular extract-transform-load pattern where each step receives the accumulated context from prior steps and appends its results.

\subsection{Step 1: Classification}\label{sec:classification}

Classification determines the type for each page in a multi-page submission. This routing decision determines which extraction pipeline and schema (for structured outputs) apply downstream.

We implement a \textbf{hybrid classification strategy} that balances cost, latency, and accuracy (Table~\ref{tab:classification}). The primary classifier uses CLIP embeddings~\citep{radford2021clip} with a k-nearest-neighbor (KNN) index trained on representative page images. CLIP-KNN runs locally with no API cost at 0.5--1\,s per page, achieving 92\% accuracy. When CLIP-KNN confidence falls below a threshold, the system falls back to a VLM classifier that sends the page image to Anthropic's Claude (Sonnet family)~\citep{anthropic2024claude} for classification.

\begin{table}[t]
\caption{Classification strategy comparison. The hybrid approach achieves near-VLM accuracy at near-CLIP cost by using VLM as a selective fallback (4\% of pages).}
\label{tab:classification}
\centering
\small
\begin{tabular}{@{}lccc@{}}
\toprule
\textbf{Strategy} & \textbf{Accuracy} & \textbf{Latency/page} & \textbf{Cost/page} \\
\midrule
CLIP-KNN & 92\% & 0.5--1\,s & \$0.000 \\
VLM & 98\% & 2--3\,s & \$0.010 \\
Hybrid & 96\% & 0.6--1.2\,s & \$0.001 \\
\bottomrule
\end{tabular}
\end{table}

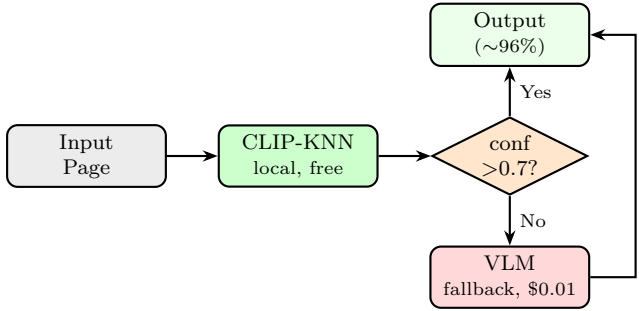
\begin{figure}[t]
\centering
\begin{tikzpicture}[
    box/.style={draw, rounded corners, align=center, minimum width=2.1cm, minimum height=0.8cm, font=\footnotesize, thick},
    decision/.style={draw, diamond, align=center, aspect=2, inner sep=1pt, font=\footnotesize, thick},
    arr/.style={-{Stealth[length=2mm]}, thick},
    lbl/.style={font=\scriptsize}
]
\node[box, fill=gray!15] (input) at (0,0) {Input\\Page};
\node[box, fill=green!20] (clip) at (2.8,0) {CLIP-KNN\\{\scriptsize local, free}};
\node[decision, fill=orange!20] (conf) at (5.6,0) {conf\\$>$0.7?};
\node[box, fill=green!8] (out) at (5.6,1.6) {Output\\{\scriptsize ($\sim$96\%)}};
\node[box, fill=red!15] (vlm) at (5.6,-1.6) {VLM\\{\scriptsize fallback, \$0.01}};

\draw[arr] (input) -- (clip);
\draw[arr] (clip) -- (conf);
\draw[arr] (conf) -- node[right, lbl] {Yes} (out);
\draw[arr] (conf) -- node[right, lbl] {No} (vlm);
\draw[arr] (vlm.east) -- ++(0.6,0) |- (out.east);
\end{tikzpicture}
\caption{Hybrid classification strategy. CLIP-KNN classifies each page locally with no API cost. When confidence exceeds 0.7 ($\sim$96\% of pages), the prediction is accepted directly. Low-confidence pages fall back to a VLM classifier at higher latency and cost.}
\label{fig:hybrid}
\end{figure}

In our tests so far, the hybrid strategy triggers VLM fallback on only 4\% of pages, reducing direct model/API cost by roughly 10$\times$ compared to VLM-only classification while recovering most of the accuracy gap. These figures exclude cluster infrastructure, storage, observability, engineering operations, and human-review labor. Figure~\ref{fig:hybrid} illustrates the decision flow. The CLIP-KNN index is stored locally and updated through MLflow model registry, enabling retraining without code changes.

\subsection{Step 2: Auxiliary Metadata Extraction}\label{sec:metadata}

For submissions requiring timeliness or provenance verification, a dedicated step extracts auxiliary metadata (e.g., date stamps, barcodes, signatures) from cover and supplementary pages. We support two backends: an RF-DETR~\citep{robinson2025rfdetr} object detection model that localizes target regions followed by recognition, and a Claude Sonnet--based extractor that processes the full page image. The object detection approach is faster but requires training data for each target format; the Claude Sonnet approach generalizes better to novel formats.

\subsection{Step 3: OCR}\label{sec:ocr}

OCR converts page images to text with word-level bounding boxes. Our primary OCR engine is DocTR~\citep{mindee2021doctr}, a PyTorch-based two-stage pipeline using a \texttt{db\_resnet50} text detection network and a \texttt{crnn\_vgg16\_bn} recognition network. DocTR processes pages at 1--2\,s per page on GPU, producing word-level text with confidence scores and bounding box coordinates.

As we discuss in \S\ref{sec:case-study}, OCR is the dominant bottleneck in the pipeline, consuming a large majority of end-to-end execution time for a typical multi-page document. This motivates the Inference Service's design: by isolating OCR inference, we can scale GPU resources specifically to address this bottleneck without over-provisioning the rest of the system.

\subsection{Step 4: Text Stitching}\label{sec:stitching}

Multi-page documents require combining OCR output from individual pages into a coherent text representation. The stitching step concatenates per-page text in page order, preserving page boundaries as metadata for downstream extraction. This step is computationally trivial ($<$1\,s) but architecturally important: it transforms the per-page OCR output into a document-level representation that the parsing step can reason over.

\subsection{Step 5: Structured Parsing}\label{sec:parsing}

The final extraction step sends the stitched OCR text to Claude Sonnet with a form-specific prompt and JSON schema. The LLM maps unstructured OCR text to structured field-value pairs, handling the semantic interpretation that rule-based extraction cannot: resolving ambiguous field references, interpreting checkbox states from OCR context, and validating internal consistency.

Each page type defines its own schema, dynamically generated from Pydantic models for type-safe validation of LLM output; the schema generator converts these to JSON Schema for inclusion in the LLM prompt.

Parsing typically consumes $\sim$4,500 input tokens (for an 8-page document's OCR text) and produces $\sim$400 output tokens of structured JSON, with a latency of $\sim$3\,s and a cost of $\sim$\$0.03 per document. Combined with hybrid classification at \$0.001 per page, the total direct API cost for an 8-page document is roughly \$0.038---approximately 80\% from parsing and 20\% from classification. While cheaper than running a VLM on raw images, this parsing cost dominates the per-document budget and motivates our decision to use OCR for text extraction rather than sending all pages directly to a VLM.

\section{A Case Study: Batch Processing at Scale}\label{sec:case-study}

To characterize the system's scaling behavior in practice, we profiled the full pipeline under controlled batch workloads of several hundred synthetic multi-page documents. We report qualitative findings here; detailed quantitative benchmarks are left to future work, pending a profiling re-run with the instrumentation improvements discussed in \S\ref{sec:lessons} (e.g., corrected stale-detection thresholds, explicit retry accounting, and decoupled queue-depth tuning).

\subsection{Single-Document Execution Profile}\label{sec:single-doc}

For a typical multi-page document, OCR dominates end-to-end wall-clock time. A representative 8-page document spends roughly two-thirds of its processing time in OCR, with LLM-based structured parsing a distant second: parsing runs once over the stitched OCR text rather than per page, so parsing latency grows sublinearly with page count---in contrast to OCR, which scales linearly because each page is processed independently. Initialization, document creation, text stitching, result upload, and database updates collectively account for a small share of execution time. Per-page OCR latency is on the order of 1--2\,s on GPU, consistent with DocTR benchmarks, and peak memory usage stays around 1\,GB, with DocTR model weights accounting for $\sim$800\,MB.

\subsection{Concurrency Saturation}\label{sec:saturation}

Under increasing concurrent load, throughput improves sharply at low concurrency and then flattens once the Inference Service's GPU capacity saturates---additional workers queue behind inference requests rather than processing in parallel. In our deployments, this inflection occurs near the ``pods $\times$ tasks per pod'' product that matches the Inference Service's steady-state request rate.

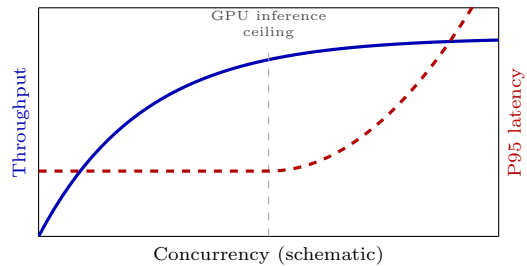
\begin{figure}[t]
\centering
\begin{tikzpicture}
\pgfplotsset{every axis/.append style={font=\scriptsize}}
\begin{axis}[
    width=0.88\columnwidth,
    height=4.6cm,
    xlabel={Concurrency (schematic)},
    ylabel={Throughput},
    axis y line*=left,
    ylabel style={blue!70!black},
    xmin=0, xmax=8,
    ymin=0, ymax=1.15,
    xtick=\empty,
    ytick=\empty,
    grid=none,
    samples=60,
    domain=0:8,
]
\addplot [blue!70!black, very thick] { 1 - exp(-0.55*x) };
\draw[gray!70, dashed, thin] (axis cs:4,0) -- (axis cs:4,0.95);
\node[gray!60!black, font=\tiny, anchor=south, align=center] at (axis cs:4,0.95) {GPU inference\\ceiling};
\end{axis}
\begin{axis}[
    width=0.88\columnwidth,
    height=4.6cm,
    ylabel={P95 latency},
    ylabel style={red!70!black},
    axis y line*=right,
    axis x line=none,
    xmin=0, xmax=8,
    ymin=0, ymax=3.5,
    ytick=\empty,
    samples=60,
    domain=0:8,
]
\addplot [red!70!black, very thick, dashed] { 1 + 0.2*(max(0, x-4))^2 };
\end{axis}
\end{tikzpicture}
\caption{Schematic saturation behavior. Throughput (solid, blue) rises at low concurrency, then flattens once the Inference Service's GPU capacity saturates. P95 per-document latency (dashed, red) stays approximately flat below the ceiling and degrades as requests queue at the Inference Service above it. Axes are illustrative; actual saturation thresholds depend on workload, model, and pod sizing.}
\label{fig:saturation-schematic}
\end{figure}

P95 per-document latency---the 95th-percentile end-to-end processing time across documents, a standard tail-latency metric in production systems---remains stable across concurrency levels below saturation, a direct benefit of the message queue mediated communication: workers wait for queue messages rather than overwhelming the Inference Service with simultaneous requests. Once concurrency exceeds the GPU inference capacity, tail latency degrades as requests queue at the Inference Service, reinforcing that the Inference Service---not orchestration---sets the saturation ceiling. Figure~\ref{fig:saturation-schematic} sketches this behavior qualitatively.

\subsection{Multi-Pod Scaling}\label{sec:multipod}

\begin{figure}[t]
\centering
\begin{tikzpicture}[
    tier/.style={draw, thick, rounded corners=2pt, align=center, font=\small\bfseries, text width=3.4cm, minimum height=0.95cm, inner sep=3pt},
    t1/.style={tier, fill=green!18},
    t2/.style={tier, fill=orange!22},
    t3/.style={tier, fill=red!20},
    lbl/.style={font=\scriptsize, align=left},
    arr/.style={-{Stealth[length=2.5mm]}, very thick}
]
\node[t1] (w)   at (0, 1.9) {Workers};
\node[t2] (inf) at (0, 0)   {Inference Service\\(GPU)};
\node[t3] (d)   at (0,-1.9) {Downstream APIs / object store};

\draw[arr] (w)   -- (inf);
\draw[arr] (inf) -- (d);

\node[lbl, anchor=west] at (2.2, 1.9)  {scales cheaply};
\node[lbl, anchor=west] at (2.2, 0)    {saturates first\\under load};
\node[lbl, anchor=west] at (2.2,-1.9)  {rate-limited /\\I/O-bound};
\end{tikzpicture}
\caption{Tiered bottleneck progression. Worker pods scale horizontally at low cost; the Inference Service's GPU capacity is the first tier to saturate as concurrency grows; adding Inference Service replicas then shifts the bottleneck to downstream services (LLM APIs, object storage) and queue overhead.}
\label{fig:bottleneck-tiers}
\end{figure}
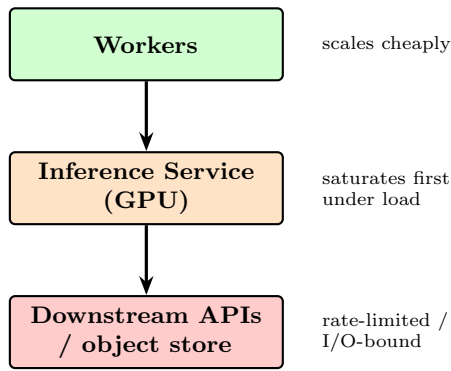

Scaling worker pods beyond the concurrency required to saturate a single Inference Service pod yields diminishing returns; the bottleneck shifts from orchestration to GPU inference. Adding Inference Service replicas unlocks further scaling, but with its own diminishing returns as the next-tier bottleneck---object-storage I/O, LLM API rate limits, or queue overhead---begins to dominate. Figure~\ref{fig:bottleneck-tiers} summarizes this tiered progression. The key architectural takeaways---that inference and orchestration must scale independently and that the Inference Service is typically the first tier to saturate---are robust across configurations. The specific break-even ratio between worker and Inference Service pods is workload-dependent and requires profiling with corrected instrumentation to state quantitatively.

One solution to the inference bottleneck might be to further decouple the inference service into services of independent operations, e.g. a classifier service, an OCR service, etc. However, we found that this approach reduced the benefits of staggering and made scaling decisions more complex, while also requiring additional engineering and devops effort. As inference is the primary cost driver for a document processing architecture, this resource is generally externally constrained and therefore a simple, horizontally scaling solution is generally the best approach.

\section{Infrastructure for Model Integration}\label{sec:infrastructure}

A production document understanding system must accommodate model evolution and improvements without system redesign. The architecture addresses this through three mechanisms.

\subsection{Containerized Inference}\label{sec:containerization}

Each OCR model is packaged as a Docker container exposing a standard REST interface. The Inference Service loads the configured model at startup and serves predictions via FastAPI endpoints. Swapping models requires changing a configuration value and redeploying the Inference Service---worker code remains untouched because it communicates through a stable API contract.

We have deployed DocTR (PyTorch, GPU), Docling (CPU-optimized), and SmolDocling (vision transformer, GPU) through this mechanism. The consistent interface allows A/B testing across models by routing a fraction of traffic to an alternate Inference Service deployment.

\subsection{Model Registry}\label{sec:registry}

Models are versioned and promoted through an MLflow model registry with environment-based aliases (\texttt{development}, \texttt{staging}, \texttt{production}). The CLIP-KNN classification index, RF-DETR detection weights, and OCR models all flow through this registry, providing reproducible deployments and rollback capability. Each model artifact is tagged with training metadata, evaluation metrics, and the dataset version used for training.

\subsection{Local vs.\ API Inference}\label{sec:local-api}

The system supports both local model inference (DocTR, CLIP, RF-DETR running on cluster GPUs) and cloud API inference (Claude Sonnet via the Anthropic API). The choice is configured per pipeline step:

\begin{itemize}[nosep]
\item \textbf{Classification}: Local CLIP-KNN with Claude Sonnet fallback.
\item \textbf{Postmark detection} (for example on envelopes for scanned correspondence): Local RF-DETR or Claude Sonnet (configurable).
\item \textbf{OCR}: Local DocTR (GPU) or Docling (CPU).
\item \textbf{Parsing}: Currently API-only (Claude Sonnet via Anthropic API).
\end{itemize}

This hybrid approach minimizes API costs for high-volume steps (classification: \$0.001/page via mostly-local inference) while leveraging cloud APIs for steps where model capability matters more than cost (parsing: \$0.03/document but only one call per document). Sensitive deployments require additional controls around provider data retention, private networking, encryption, audit logging, and retention policies; in stricter environments, the same service contract can be backed by self-hosted models or private model endpoints rather than public APIs.

\subsection{Observability}\label{sec:observability}

Production operation depends on observability at the document, step, and service levels. At minimum, the system must expose document-level status transitions, per-step latency, queue depth, retry counts, and failure attribution across the Gateway, Worker, and Inference Service tiers. These signals serve different purposes: queue depth and worker concurrency reveal whether orchestration is keeping pace with ingress; inference latency and GPU saturation reveal when model-serving capacity is the active bottleneck; document-level statuses and structured error codes reveal whether failures are concentrated in OCR, parsing, object-storage access, or downstream API dependencies.

This distinction matters because many symptoms look similar from the outside. A growing queue might indicate underprovisioned workers, an inference bottleneck, or repeated retries caused by an unreliable downstream dependency. Likewise, a rise in end-to-end document latency might reflect OCR slowdown, API throttling, or queue contention rather than a regression in the final parsing step. The architecture therefore benefits from observability aligned to service boundaries and pipeline steps, not just coarse application-level success rates.

\section{Lessons Learned}\label{sec:lessons}

Operating this architecture in production revealed several failure modes and design insights not apparent from model benchmarks alone.

\paragraph{Message queue visibility timeout must match processing time.}
Our initial visibility timeout of 30\,s caused documents to be re-delivered to other workers while still being processed (typical processing takes 15--25\,s per document). This produced duplicate results and wasted compute. Setting the visibility timeout to 300\,s---well above the P99 processing time---eliminated re-delivery without meaningfully delaying failure detection.

\paragraph{OCR is the bottleneck, not the LLM.}
Intuition suggested that LLM parsing would dominate latency and cost. In practice, OCR dominates end-to-end execution time while LLM parsing takes a much smaller share. This is because OCR processes every page independently (e.g., 8 sequential inference calls for an 8-page document), while parsing processes the full document in a single LLM call. Optimizing OCR throughput---through batched inference, model distillation, or faster architectures---yields greater system-level improvements than optimizing the LLM step.

\paragraph{Model accuracy $\neq$ system reliability.}
A VLM classifier achieving 98\% accuracy in evaluation still produces 2\% misclassifications in production, routing documents to incorrect extraction pipelines. At 1,000 documents/day, this means 20 daily failures requiring human review. The hybrid classification strategy mitigates this challenge by using CLIP-KNN's different failure distribution as a complementary signal, but the lesson generalizes: system reliability requires defense in depth, not just high model accuracy.

\paragraph{Stale detection is harder than it sounds.}
Workers must detect when a document has been processing for too long and is likely stuck. Our initial implementation tracked time from when the document entered the worker's local queue, not from when inference actually started. Documents waiting in the local queue for inference capacity appeared ``stale'' and were incorrectly marked as failed. The fix---tracking \texttt{processing\_start\_time} from the first Inference Service call---required careful state management across async tasks.

\paragraph{Changing models requires scaling reanalysis.}
While staggering is a useful technique for us to improve resource utilization, it also required us to consider the implications of adding new models to the pipeline. Generally our experimentation focused on improving accuracy or model performance, but we found that we needed to additionally consider the impact on the worker-to-inference ratio and the potential for resource contention. Generally speaking, given two models that are approximately equivalent in accuracy, the model that uses different computational resources than the previous or next step in the pipeline is preferable even if the throughput is lower on a per-document basis.

\paragraph{Limitations of the present case study.}
The empirical discussion in this paper is intentionally scoped as an operational case study rather than a fully controlled systems benchmark. Saturation points, worker-to-inference ratios, and per-step cost tradeoffs depend on workload composition, page count distribution, document quality, queue settings, model choice, and the behavior of downstream managed APIs. Our observations therefore support the architectural claims qualitatively and directionally, but they should not be treated as universal constants. Teams adopting similar designs should re-profile with their own workloads, hardware, and retry policies before transferring the exact thresholds reported here.

\section{Conclusion}\label{sec:conclusion}

We have described a production architecture for document understanding that processes structured forms through a queue-driven pipeline of classification, OCR, and language-model-based extraction. The system's key architectural decisions---separating inference from orchestration, using hybrid classification to balance cost and accuracy, and communicating through message queues for decoupled scaling---reflect tradeoffs that we believe generalize beyond our specific deployment.

The qualitative profiling observations presented here are intended to help practitioners make informed infrastructure decisions before investing in a full benchmark suite. Our finding that OCR, not LLM parsing, dominates end-to-end latency may surprise teams planning document understanding deployments, and motivates investment in OCR optimization or alternative architectures that bypass OCR entirely.

Looking ahead, we see three directions for further evolution of this class of systems. First, end-to-end VLMs that can process page images directly may eventually replace the OCR step, simplifying the pipeline at the cost of higher per-page inference cost. Second, retrieval-augmented approaches~\citep{faysse2024colpali} could enable selective processing of only relevant pages from large document collections, reducing total compute. Third, efficiency-optimized models may shift the cost-accuracy Pareto frontier enough to make VLM-only pipelines economically viable at scale.

The gap between model research and production deployment remains wide. We hope that describing this architecture, its operating characteristics, and the lessons we learned running it contributes to closing that gap.

\bibliographystyle{plainnat}
\bibliography{references}

\end{document}